\documentclass[journal, letterpaper]{IEEEtran}
\usepackage{graphicx}
\usepackage{url}        
\usepackage{amsmath}    
\usepackage{textgreek}	% Greek to me, dawg
\usepackage{listings}
\usepackage{csvsimple}
\usepackage{longtable}

\usepackage{subcaption}
\usepackage{pgfplotstable}
\usepackage{tikz}
\usepackage[flushleft]{threeparttable}
\usepackage{float}
\usepackage{multirow}
\begin{document}

\title{Real-Time High-Quality Stereo Matching System on a GPU}
\author{Qiong Chang, Tsutomu Maruyama}
\maketitle

\begin{abstract}
In this paper, we propose a low error rate and real-time stereo vision system on
GPU. Many stereo vision systems on GPU have been proposed to date. In those
systems, the error rates and the processing speed are in trade-off
relationship. We propose a real-time stereo vision system on GPU for the high
resolution images. This system also maintains a low error rate compared to other
fast systems. In our approach, we have implemented the cost aggregation (CA),
cross-checking and median filter on GPU in order to realize the real-time
processing. Its processing speed is 40 fps for $1436\times{992}$ pixels images
when the maximum disparity is 145, and its error rate is the lowest among the
GPU systems which are faster than 30 fps.
\end{abstract}

\section{Introduction}
The aim of stereo vision systems is to reconstruct the 3-D geometry of a scene
from images taken by two separate cameras. The computational complexity of the
stereo vision is very high, and many acceleration systems with GPUs, FPGAs and
dedicated hardware have been developed
\cite{Global_FPGA}\cite{Self-distributed}. All
of them succeeded in real-time processing of high resolution images, but their
accuracy is not good enough because they simplify the algorithms to fit the
hardware architecture.  In this paper, we aim to construct a real-time GPU
stereo vision system for the high resolution image set
($1436\times{992}\times{145}$ disparities) in the Middlebury Benchmark.

GPUs have more than hundred cores which run faster than 1GHz, and drastic
performance gain can be expected in many applications. However, the processing
speed of the stereo vision by GPUs is much slower than FPGAs such as
\cite{Fuzzy_logic}\cite{End_end_GPU}. This is caused by the fact that data for
several lines are intensively accessed at a time in the stereo vision, but the
shared memory is too small to cache those lines. Therefore, many memory accesses
to the global memory are required, and they limit the processing speed by
GPUs. On the other hand, more sophisticated algorithms can be implemented on
GPUs than FPGAs, and lower error rates have been achieved by many GPU
systems\cite{MC-CNN-fst}\cite{FEN-D2DRR}. The algorithms for the stereo vision
can be categorized into two groups: local and global. In the local algorithms,
only the local information around the target pixel is used to decide the
disparity of the pixel, while the disparities of all pixels are decided
considering the mutual effect of all pixels in the global algorithms. Thus, in
general, the global algorithms achieve lower error rates, but require longer
computation time. On many GPUs, algorithms which require only local information
in each step, but propagate the mutual effect gradually by repeating the step
many times are implemented. Their error rates are low enough, but their
processing speed is far behind of the real-time requirement.

\begin{figure}
 \centering
  \includegraphics[width=3in]{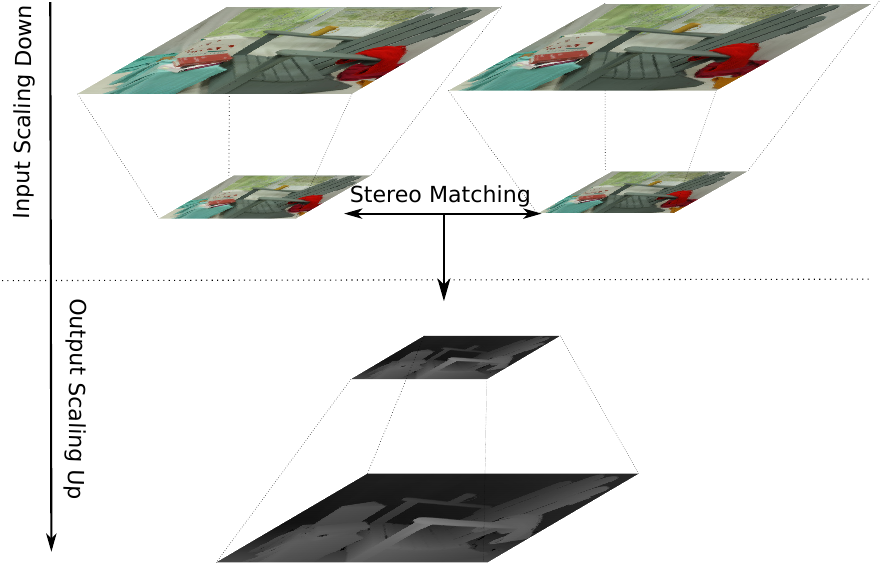}
    \caption{Image Scaling}
  \label{fig:scale}
\end{figure}

We have implemented a local search algorithm, which is an improved
version of the algorithm that we have implemented on
FPGA\cite{Fast_FPGA}. In our algorithm, AD (absolute difference) and
mini-Census transform \cite{Mini-census} are used to calculate the
matching costs of the pixels in the two images, and they are aggregated
along the $x$- and $y$-axes by using the Cross-based method
\cite{cross-based1}\cite{cross-based2}, in order to compare the pixels
as a block of the similar color. The matching costs are calculated twice
using left and right image as the base, and two disparity maps are
generated. Then, the disparity map is improved using the disparities
which are common in both disparity maps. These operations are chosen to
realize low error rates without repetitive computation. However, when we
consider processing a high resolution image, its computational
complexity is too high to achieve real-time processing. In order to
achieve high performance, in this paper, we scale down the images to
reduce the computational complexity. As shown in Fig.\ref{fig:scale}, to
reduce the computational complexity, we scale down the input images into
a small size, and then scale up the disparity map to the original
size. The images are scaled down to 1/4 by reducing the width and height
by half, and the disparities are calculated on the scaled down
images. The maximum disparity is also reduced to half, which means that
the total computational complexity can be reduced to 1/8. This approach
typically worsens the matching accuracy, but in our approach, an
bilateral interpolation method is performed during the scaling up step,
and a high matching accuracy can be maintained. This approach becomes
possible because of the high quality of the high resolution images.

\section{STEREO VISION}
\begin{figure}
 \centering
  \includegraphics[width=3in]{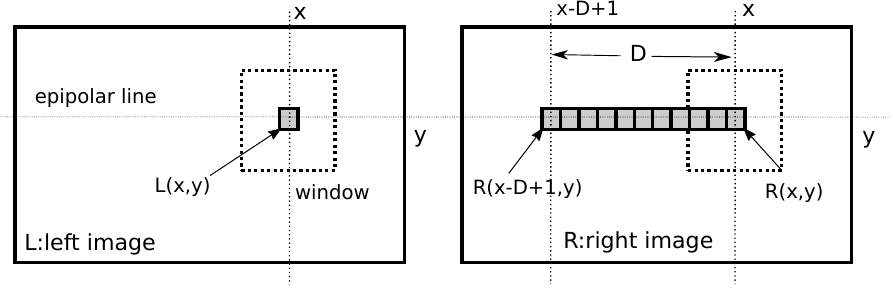}
    \caption{Local matching under the epipolar restriction}
  \label{fig:stereo}
\end{figure}

In the stereo vision systems, the matching of pixels in the two images (left and
right) taken by two separate cameras is searched to reconstruct the 3-D geometry
of the scene. When the two cameras are calibrated properly, epipolar restriction
can be used. Under this restriction, we can obtain a disparity map $D_{map}$ by
finding the matching of pixels on the epipolar lines of the two images as shown
in Fig.\ref{fig:stereo}. Fig.\ref{fig:stereo} shows how to calculate a disparity
map using the left image as the base. A pixel in the left image $L(x,y)$ (or a
window centered by $L(x,y)$) is compared with $D$ pixels $R(x-d,y) d=[0,D-1]$ in
the right image (or $D$ windows centered by those pixels), and the most similar
pixel to $L(x,y)$ is searched. $D$ is the maximum disparity value. Suppose that
$R(x-d,y)$ is the most similar to $L(x,y)$. Then, this means that $L(x,y)$ and
$R(x-d,y)$ are the same point of an object, and the distance $Z$ to the object
can be calculated from d and two parameters of the two cameras using the
following equation:
\begin{equation}
%Z=f\cdot{\frac{B}{d}}.
Z=f{\frac{B}{d}}.
\end{equation}
Here, $f$ is the focus length of both cameras, and $B$ is the distance between
the two cameras. Smaller d means that the object is farther away from the
cameras, and $d=0$ means that the object is at infinity. The main problem in the
stereo vision is to find the matching pixels in the two images correctly.

Another important problem in the stereo vision is the occlusion. When an
object is taken by two cameras, some parts of the object appear in one
image but do not appear in another image, depending on the positions and
angles between the cameras and the object. These occlusions are the
major source of errors in the stereo vision systems. In order to avoid
those errors, $D_{map}$ is calculated twice: once using $L$, the left
image, as the base ($D^L_{map}$), and another using $R$, the right image, as
the base ($D^R_{map}$). Then, by using the same matching found in both
$D^L_{map}$ and $D^R_{map}$ as the reliable disparities (called the
ground control points, or GCPs), higher quality disparity maps can be
obtained.

\section{OUR ALGORITHM}
In this section, we introduce the details of our algorithm. Our algorithm
consists of the following steps.
\begin{enumerate}
\item the two input images are gray-scaled
\item scaling down the two images
\item calculating matching cost of each pixel
\item cost aggregation along the $x$- and $y$-axes
\item generating two disparity maps
\item detecting GCPs (ground control pixels) by cross-checking the two
  disparity maps
\item refinement by a median filter and filling the non-GCPs by using a
  bilateral estimation method
\item scaling up the disparity map
\end{enumerate}

\subsection{Scaling Down}
In order to reduce the computational complexity, the two images are scaled down
linearly in both horizontal and vertical directions using the mean-pooling
method. Here, take the left image as an example:
\begin{align}
&L(x,y)=\nonumber\\
&\frac{1}{(2m+1)^2}\times{\sum_{j=-m}^{m}\sum_{i=-m}^{m}L_{org}(K\cdot{x}+i, K\cdot{y}+j)}
\end{align}
where $L_{org}(K\cdot{x}+i, K\cdot{y}+j)$ is the pixel in the original
image, and $K$ is the factor for the scaling down (in our
implementation, $K=2$). $L(x,y)$ is the pixel of the left scaled down
image, and is smoothed by a mean-filter the size of which is $(2m +
1)^2$. By choosing the block size carefully, we can avoid the loss of
the matching accuracy, and can improve the processing speed.

\subsection{Matching cost between two pixels}
\begin{figure}
 \centering
  \includegraphics[width=3in]{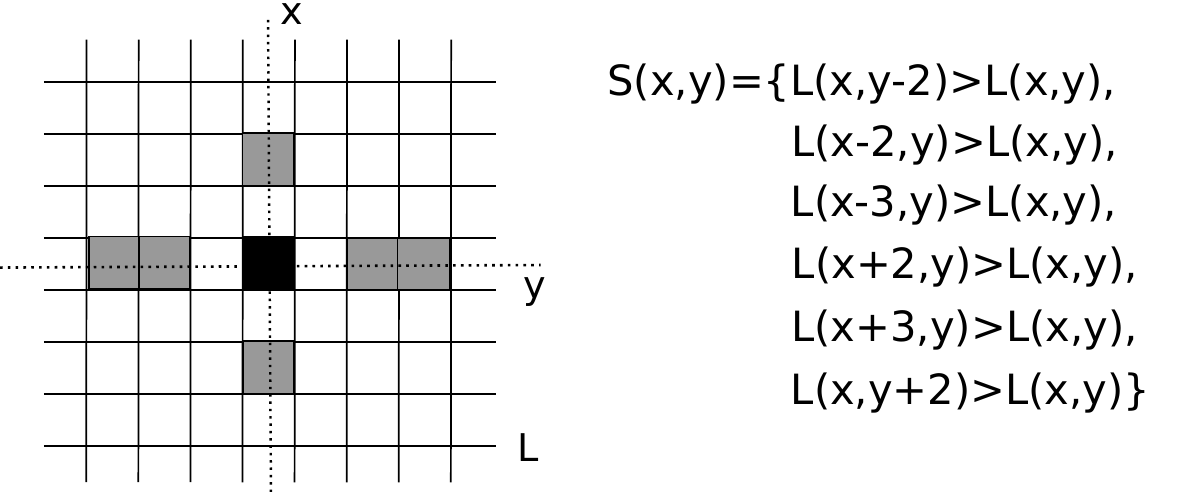}
    \caption{Mini-census Transform}
  \label{fig:census}
\end{figure}

The matching cost of each pixel is calculated using the absolute difference of
the brightness and the mini-census transform.  When the left image $L(x,y)$ is
used as the base, the matching cost of the disparity $= d$ is given by
\begin{equation}
C^L(x,y,d)=C^L_{AD}(x,y,d)+C^L_{MC}(x,y,d).
\end{equation}
$C^L_{AD}(x,y,d)$ is the cost by the absolute difference of the brightness of the
two pixels, and given by 
\begin{equation}
C^L_{AD}(x,y,d)=1-\exp(-\frac{|L(x,y)-R(x-d,y)|}{\lambda_{AD}}) 
\end{equation}
where $\lambda_{AD}$ is a constant, In the same way, $C^L_{MC}(x,y,d)$, the cost
by mini-census transform, is given by
\begin{equation}
C^L_{MC}(x,y,d)=1-\exp(-\frac{MC(L(x,y), R(x-d,y))}{\lambda_{MC}})
\end{equation}
where $\lambda_{MC}$ is a constant, and MC($\alpha$, $\beta$) is the Hamming
distance between the mini-census transform of $\alpha$ and $\beta$. Mini-census
transform used in our approach is shown in Fig.\ref{fig:census}. The center
pixel $L(x,y)$ is compared with its six neighbors, and a six bit sequence is
generated as shown in Fig.\ref{fig:census}. This approach is based on the
hypothesis that the relative values of the brightness are kept in both images.

When the right image $R(x,y)$ is used as the base, the matching cost is given as
follows.
\begin{equation}
\begin{split}
C^R(x,y,d)&=C^R_{AD}(x,y,d)+C^R_{MC}(x,y,d)\\
&=1-\exp(-\frac{|R(x,y)-L(x+d,y)|}{\lambda_{AD}})\hspace{0.5em}+ \\
&\hspace{1.3em} 1-\exp(-\frac{MC(R(x,y), L(x+d,y))}{\lambda_{MC}})\\
&=C^L(x+d,y,d).
\end{split}
\end{equation}

This equation means that all $C^R(x,y,d)$ are already calculated when
$C^L(x,y,d)$ are calculated, and $C^L(x,y,d)$ can be reused as
$C^R(x-d,y,d)$.

\begin{figure}
 \centering
  \includegraphics[width=3in]{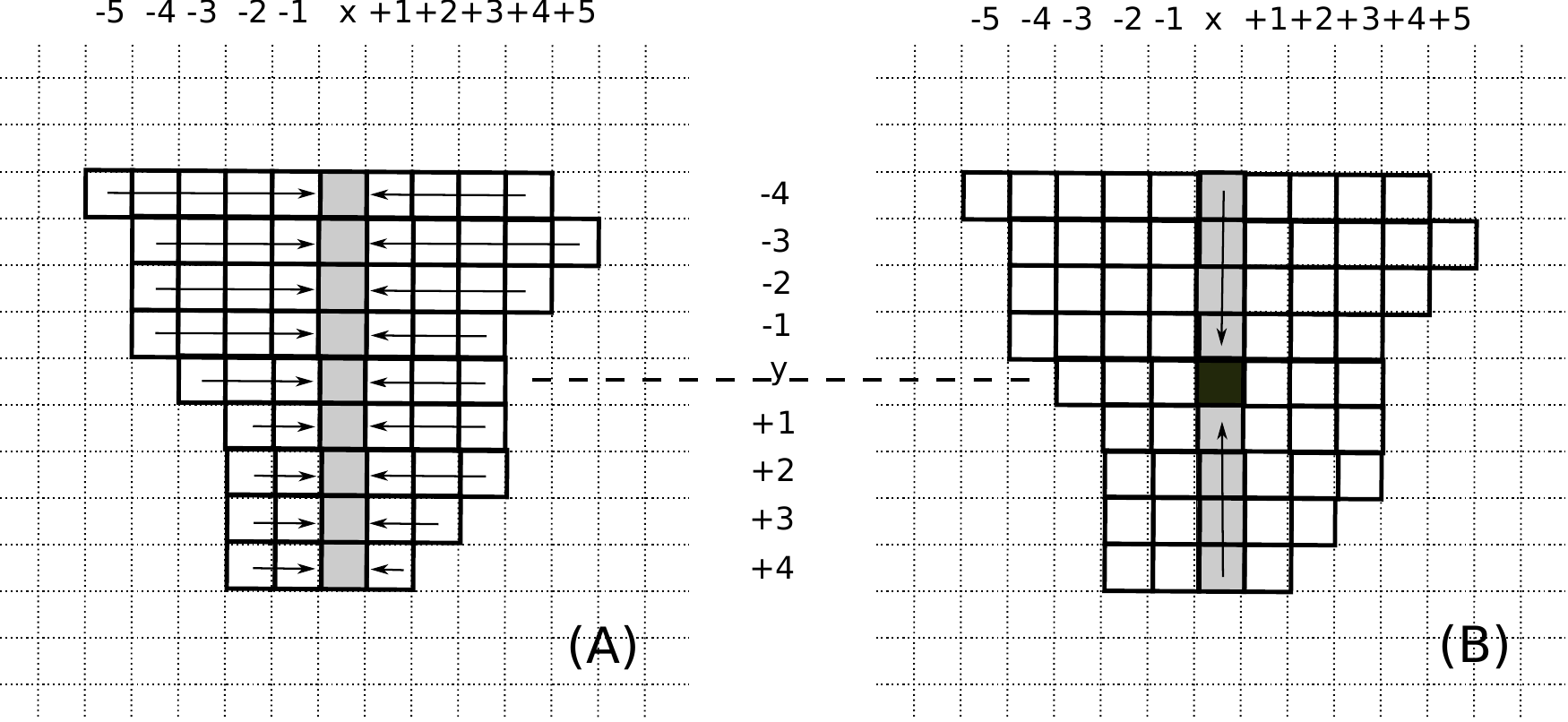}
    \caption{A cost aggregation method}
  \label{fig:CA}
\end{figure}

\subsection{Cost Aggregation}
The matching costs are aggregated as much as possible considering the similarity
of the brightness of the pixels to compare the pixels as a block of the similar
brightness. Fig.\ref{fig:CA} shows how the matching costs are aggregated. First,
the matching costs are aggregated along the $x$-axis.
\begin{equation}
CA^L_x(x,y,d)=\sum^{+n}_{dx=-m}C^L(x+dx,y,d)
\end{equation}
Here, m and n are the number of the continuous pixels with the similar
brightness to $L(x,y)$ $(|L(x,y)-L(x+dx,y)|<\delta)$ on the left and right-side
of $L(x,y)$. For example, in Fig.\ref{fig:CA}, $m=3$ and $n=3$, because all pixels from
$L(x-3,y)$ to $L(x+3,y)$ are similar to $L(x,y)$. Then, $CA^L_x(x,y,d)$ are
aggregated along the $y$-axis as 
\begin{equation}
CA^L(x,y,d)=\sum^{+N}_{dy=-M}CA^L_x(x,y+dy,d).
\end{equation}
Here, $M$ and $N$ are the number of the continuous pixels with similar
brightness to $L(x,y)$ on the upper and lower side of $L(x,y)$. In
Fig.\ref{fig:CA}, $M$ is 4 and $N$ is 4, because the pixels from $L(x,y-4)$ to
$L(x,y+4)$ are similar to $L(x,y)$.

Then, $d$ which minimizes $CA^L(x,y,d)$ is chosen as the disparity at
$L(x,y)$, and disparity map $D^L_{map}(x,y)$ is obtained.
\begin{equation}
D^L_{map}(x,y)=\min_d{CA^L(x,y,d)}.
\end{equation}
By enlarging the range for summing up along the $x$- and $y$- axes, we can obtain
more accurate disparities, though it requires more computation time.

\subsection{GCPs}
In our approach, $D^R_{map}(x,y)$ is also calculated in the same way as
$D^L_{map}(x,y)$. Then, ground control points, or GCPs, are obtained by
comparing them \cite{LargeStereo}. Suppose that $D^L_{map}(x,y)=k$. This means
that $L(x,y)$ and $R(x-k,y)$ showed the best matching when the left image is
used as the base, and they are the same point of the object in the images.
Therefore, $D^R_{map}(x-k,y)$ should also be $k$. If this requirement
\begin{equation}
D^L_{map}(x,y)=D^R_{map}(x-k,y)=k
\end{equation}
is satisfied, the point is called a GCP, and it is considered that GCPs have
higher reliability.

\subsection{Bilateral Estimation}
\begin{figure}
\centering
  \includegraphics[width=3in]{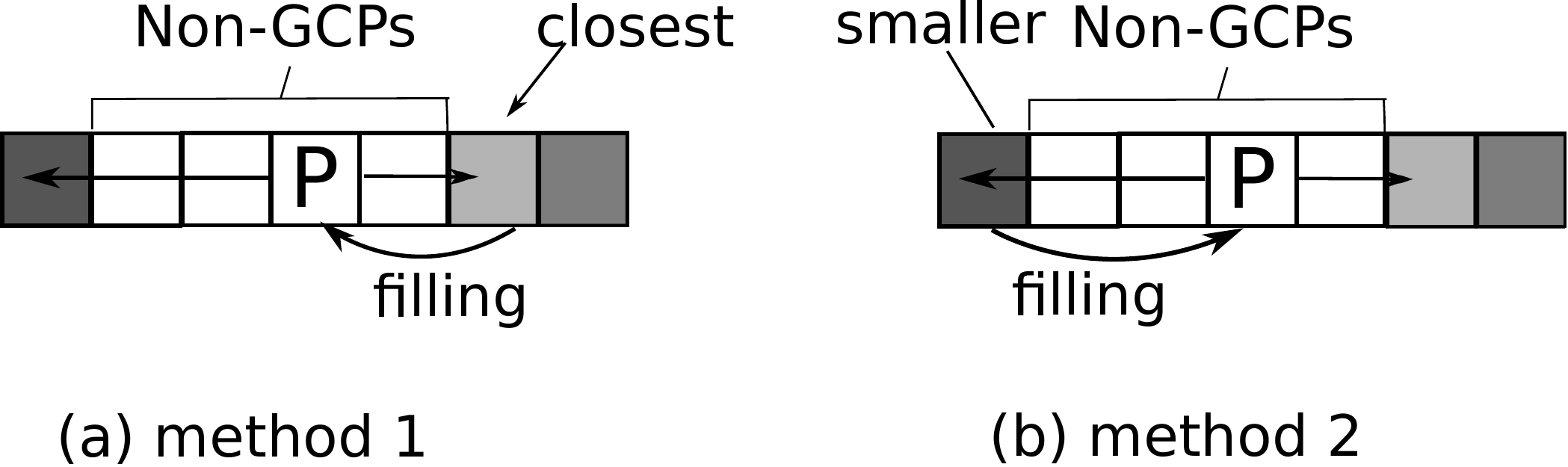}
    \caption{Non-GCPs Filling}
  \label{fig:GCPs_Filling}
\end{figure}

Ideally, all pixels except for those in the occluded regions should be GCPs,
however, in actuality more pixels become non-GCPs because of the slight change
of the brightness between the input images. To achieve more reliable disparities
of non-GCPs, two approaches are often used \cite{Sub-pixel}. In both approaches,
for each non-GCP, the closest GCPs on the left and right hand-side along the
$x$-axis are searched first. Then, in the first approach, as shown in
Fig.\ref{fig:GCPs_Filling}(a), the closer GCP in the distance is chosen as the
disparity of the non-GCP because the non-GCP and the closer GCP can be
considered to belong to the same object with a higher probability. In the second
approach, as shown in Fig.\ref{fig:GCPs_Filling}(b), the smaller disparity is
chosen as the disparity of the non-GCP assuming that the non-GCP is caused by
the occlusion. The disparity of the occluded region is smaller than that of the
foreground object because the disparity of the closer object is larger, and the
non-GCP should have a smaller disparity. Both of these two methods can be easily
implemented on GPU because of their high parallelism. However, due to their
single function, the overall accuracy is not good enough.

In our system, we proposed a bilateral estimation methods to fill the non-GCPs
as following steps:
\begin{enumerate}
  \item Define the disparities of the GCP of $L(x-i,y)$ and $L(x+j,y)$ as $D(x-i,y)$ and $D(x+j,y)$.
  \item If $|D(x-i,y)-D(x+j,y)|\le{T}$, where $T$ is the threshold for the
    difference of disparity, it can be considered that the disparity is changing
    continuously in this range, and $D(x,y)$ is filled as:
    \begin{equation}
    D(x-i,y)+i\cdot((D(x-i,y)-D(x+j,y))/(i+j)).
    \end{equation}
  \item If $|D(x-i,y)-D(x+j,y)|>T$, which means that the disparity changes
    rapidly in this range, it can be considered that an edge exists in this
    range. Thus $D(x,y)$ is chosen as the $D(x-i,y)$ if $L(x-i,y)$ is closer to
    $L(x,y)$ than $L(x+j,y)$ in color, and otherwise, $D(x,y)$ is chosen as
    $D(x+j,y)$.
\end{enumerate}
With this approach, we can fill the different areas in the different
methods. Then, an accurate disparity map can be expected.

\section{IMPLEMENTATION ON GPU}
We have implemented the algorithm on Nvidia GTX780Ti.
GTX780Ti has 15 streaming multi-processors (SMs).
Each SM runs in parallel using 192 cores in it (2880 cores in total).
GTX780Ti has two-layered memory system. Each SM has a 48KB shared
memory, and one large global memory is shared among the SMs. The access
delay to the shared memory is very short, but that to the global memory
is very long. Therefore, the most important technique to achieve high
performance on GPU is how to cache a part of the data on the shared
memory, and to hide the memory access delay to the global memory. The
shared and the global memory have the restriction of the access to
them. In the CUDA, which is an abstracted architecture of Nvidia's GPUs,
16 threads are managed as a set. When accessing to the global memory, 16
words can be accessed in parallel if the 16 threads access to continuous
16 words which start from 16 word-boundary.  Otherwise, the bank
conflict happens, and several accesses to the global memory happens. The
shared memory consists of 16 banks, and in this case, the 16 words can
be accessed in parallel if they are stored in the different memory banks
(the addresses of the 16 words do not need to be continuous). For
reducing the memory accesses to the global memory, the order of the
calculation on GPU is different from the one described in the previous
section.

\begin{itemize}
\item \textbf{Step1} transfer the input images onto the GPU and scale down them
\item \textbf{Step2} compare the brightness of the pixels along the $x$-axis
\item \textbf{Step3} calculate the matching costs and aggregate them along the $x$-axis
\item \textbf{Step4} compare the brightness of the pixels along the $y$-axis
\item \textbf{Step5} aggregate the cost along the $y$-axis, and generate two disparity maps
\item \textbf{Step6} find GCPs by cross-checking
\item \textbf{Step7} apply median filter to remove noises and estimate the
  disparity map using bilateral method
\item \textbf{Step8} scale up the disparity map and transfer back to the CPU.
\end{itemize}

\begin{figure}
\centering
  \includegraphics[width=3in]{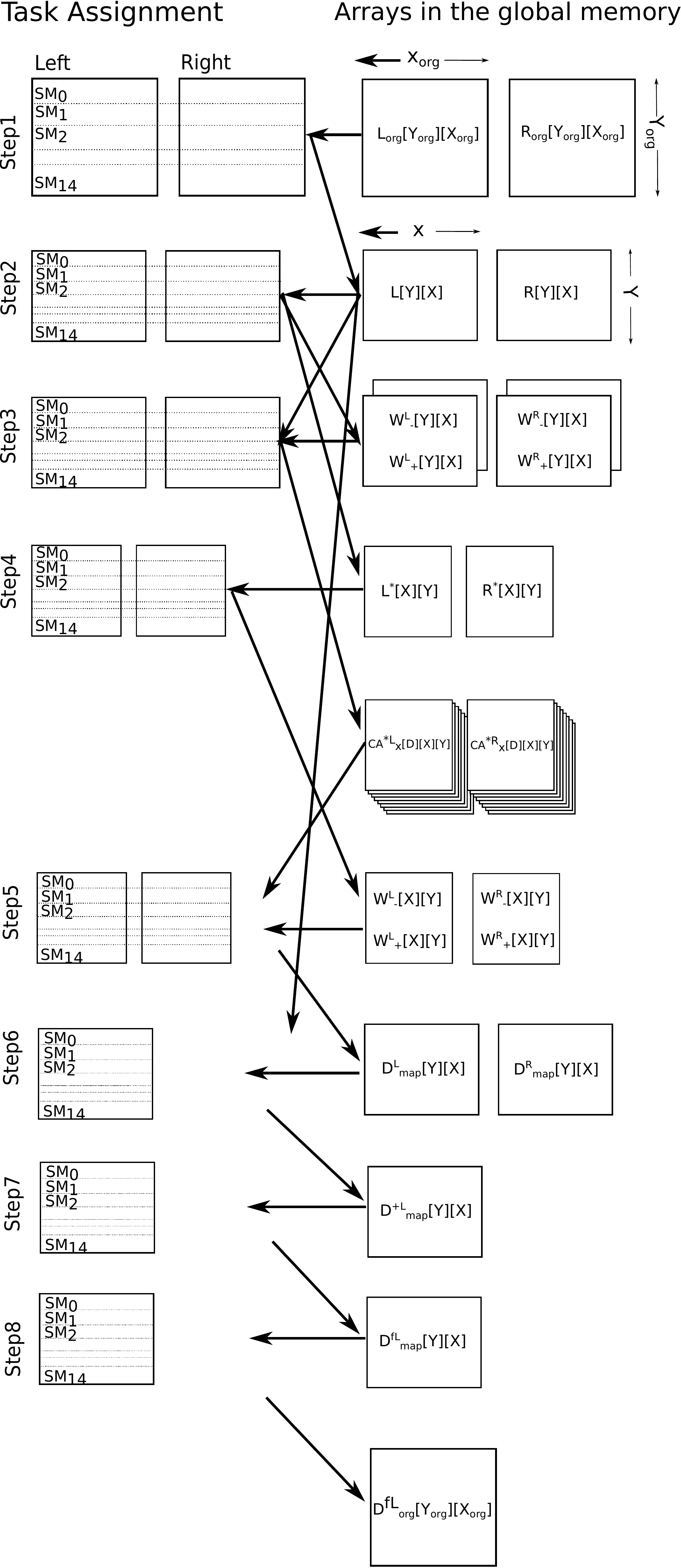}
    \caption{Task assignment of each step}
  \label{fig:pipeline}
\end{figure}

In the following discussion, $X_{org}\times{Y_{org}}$ is the image size
($X_{org}$ is the width, and $Y_{org}$ is the height), and $L_{org}[y][x]$ and
$R_{org}[y][x]$ are the pixels in the left and right images. $L[y][x]$ and
$R[y][x]$ are the pixels in the scaled-down images, and $X\times{Y}$ is the
image size of them ($X=X_{org}/2$, $Y=Y_{org}/2$). Fig.\ref{fig:pipeline} shows
the task assignment and input/output of each step. The details are discussed in
the following subsections.
\subsection{Step1}
\begin{figure}
\centering
  \includegraphics[width=3in]{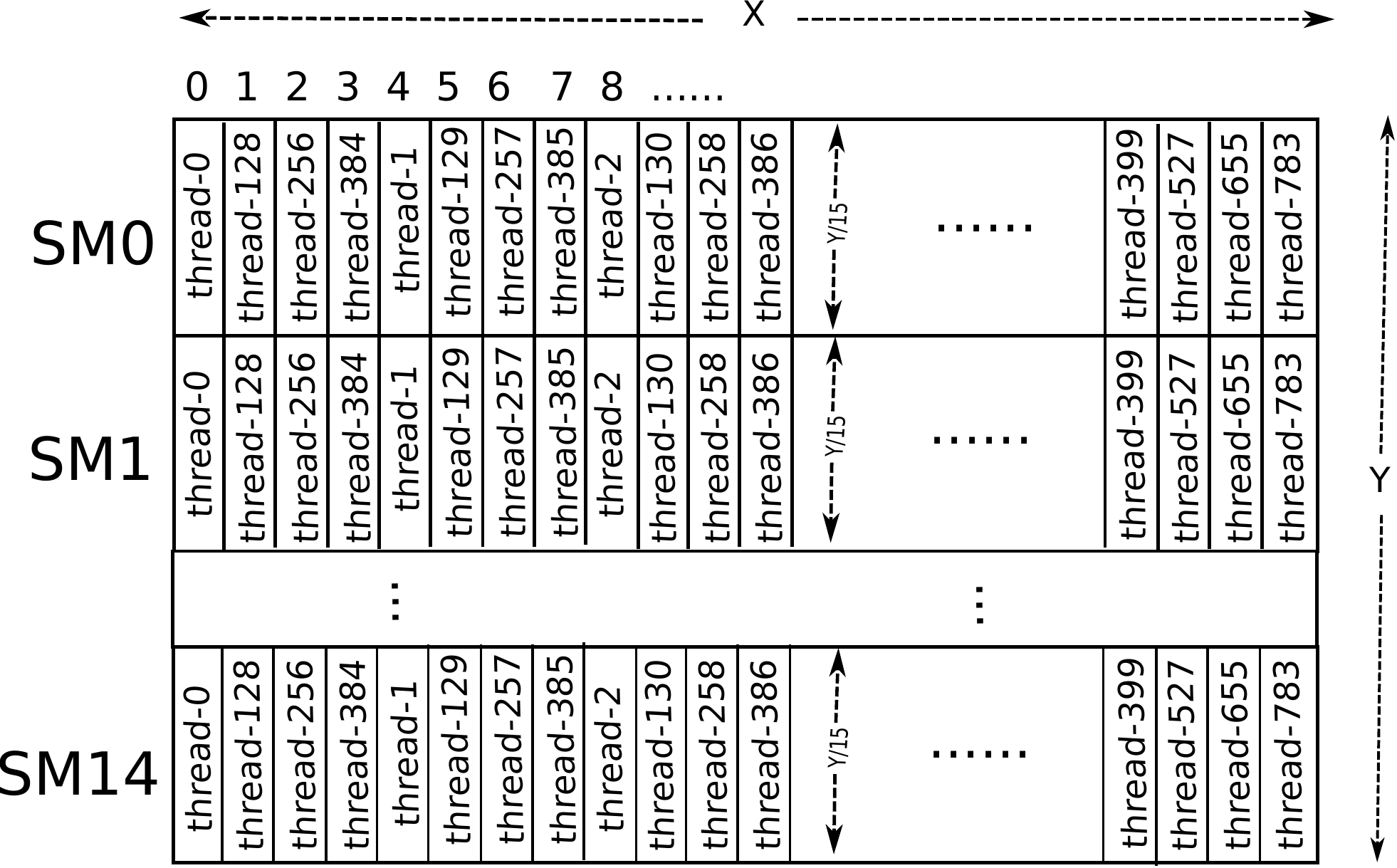}
    \caption{Mapping pixels to the threads}
  \label{fig:mapping}
\end{figure}

The inputs to this step are $L_{org}[y][x]$ and $R_{org}[y][x]$, and
they are transferred onto the global memory of the GPU, and are
processed in parallel using 15 SMs.
\begin{enumerate}
\item $Y_{org}/15$ lines of both images are assigned to each SM as shown
  in Fig.\ref{fig:mapping}.
\item $X_{org}$ columns in the $Y_{org}/15$ lines are processed using
  $X'$ ($X_{org}\le{X'}$) threads in the SM ($X'$ must be a multiple of
  64 because of the reason described below).  When $X'>X_{org}$,
  $X'-X_{org}$ threads work in the same way as the $X_{org}$ threads,
  but generate no outputs.
\item When $X_{org}$ is larger than the maximum number of the threads in
  one SM (1024), each thread processes more than one columns. In our
  implementation, because the resolution of the input images are greater
  than 1024, each thread processes 2 columns during the scaling-down
  step.
  %Thus, the $X'$ need to be larger than $X_{org}/2$. ????
\end{enumerate}
For each pixel $L_{org}(x_{even},y_{even})$, both of the vertical and
horizontal coordinates of which are even, all of the surrounding pixels
$L_{org}(x_{even}+dx,y_{even}+dy)$ ($dx\in[-1,1]$, $dy\in[-1,1]$) are
added together. Then, as the pixel value of the scaled-image, the
average of the summation is stored in the global memory.
\subsection{Step2}
For $Y/15$ pixels in one column (let the pixels be
$L[y_b+k][x_b]$($k=0,14$)), each thread compares its pixel's value with
its neighbors along the $x$-axis ($L[y_b+k][x_b+dx]$ ($dx=1,W_x$) and
$L[y_b+k][x_b-dx]$ ($dx=1, W_x$)). The data type of $L[y][x]$ is $8b$
(unsigned char). Therefore, four continuous pixels are packed in one
$32b$ word, and stored in the same memory bank of the shared
memory. This means that these four continuous pixels can not be accessed
in parallel owing to the memory access restriction of the shared
memory. In order to avoid the bank conflict, {\it Thread}$_i$ processes column
($i/(X/4)+(i\times{4})\%X$) as shown in Fig.\ref{fig:mapping}. In
Fig.\ref{fig:mapping}, the first four pixels of each line ($L[y][0]$,
$L[y][1]$, $L[y][2]$, $L[y][3]$) are stored in the first bank, and next
four pixels ($L[y][4]$, $L[y][5]$, $L[y][6]$, $L[y][7]$) in the second
bank. {\it Thread}$_0$ processes $Y/15$ pixels on $x=0$
($L[y_b+dy][0]$ ($dy=0,14$)) sequentially, and {\it Thread}$_1$ processes
$Y/15$ pixels on $x=4$ ($L[y_b+dy][4]$ ($dy=0,14$)) sequentially. By
changing the order of the computation like this, bank conflict can be
avoided. In our algorithm, all pixels can be processed independently,
and the same results can be obtained regardless of the computation
order.

In this method, four continuous pixels are stored in the same bank, and 16
threads are executed at the same time in CUDA. Therefore, $X$ must be a multiple
of 64 ($4\times{16}$).  When, $X$ is not the multiple of 64, larger $X$ which is
the multiple of 64 is chosen, and the computation results for the extended part
are discarded.

The outputs of this step are two integer values for each pixel, which
show how many pixels are similar to the center pixel to the plus/minus
direction of the $x$-axis. These values for $L[y][x]$ are stored in
$W^L_-[y][x]$ and $W^L_+[y][x]$, and those for $R[y][x]$ are stored in
$W^R_-[y][x]$ and $W^R_+[y][x]$. The data width of these arrays is
$32b$, and the direct access to these values causes no bank
conflict. $L[y][x]$ and $R[y][x]$ are transposed here, and stored in
$L^{*}[x][y]$ and $R^{*}[x][y]$ respectively.

\subsection{Step3}
In this step, first, two matching costs ($C^L(x,y,d)$ and $C^R(x,y,d)$) are
calculated, and then, they are aggregated along the $x$-axis using the range
information in $W^L_-[y][x]$, $W^L_+L[y][x]$, $W^R_{-}[y][x]$ and
$W^R_{+}[y][x]$ to calculate $CA^L_x(x,y,d)$ and $CA^R_x(x,y,d)$. Here,
actually, we do not need to calculate $C^R(x,y,d)$ as described in Section
3.C because $C^L(x+d,y,d)$ can be used as $C^R(x,y,d)$. Therefore, all SMs
are used to calculate $C^L(x,y,d)$ as shown in Fig.\ref{fig:pipeline}-$step3$,
and each SM processes $Y/15$ lines as follows.
\begin{enumerate}
\item For each of the Y/15 lines, repeat the following steps.
\item Set $d=0$.
\item Calculate $C^L(x,y,d)$ for all $x$ in the current line. $C^L(x,y,d)$ is
  stored in $C[x]$ (an array in the shared memory). For this calculation, 3
  lines of $L[y][x]$ and $R[y][x]$ are cached in the shared memory for
  calculating the mini-census transform, and they are gradually replaced by the
  next line as the calculation progresses.
\item Calculate $CA^L_x(x,y,d)$ as follows.
  \begin{enumerate}
        \item Set $CA_x[x]=C[x]$.  
        \item Add $C[x+dx]$ to $CA_x[x]$ starting from $dx=1$ to the
          position given by $W^L_+[y][x]$.  
        \item Add $C[x-dx]$ to $CA_x[x]$ starting from $dx=1$ to the
          position given by $W^L_-[y][x]$.
        \item Store $CA_x[x]$ into $CA^{*L}_{x}[d][x][y]$ in the global
          memory (note that this array is transposed).
  \end{enumerate}
\item Calculate $CA^R_x(x,y,d)$ as follows.
  \begin{enumerate}
        \item Set $CA_x[x]=C[x+d]$.  
        \item Add $C[x+d+dx]$ to $CA_x[x]$ starting from $dx=1$ to the
          position given by $W^R_+[y][x]$.  
        \item Add $C[x+d-dx]$ to $CA_x[x]$ starting from $dx=1$ to the
          position given by $W^R_-[y][x]$.
        \item Store $CA_x[x]$ into $CA^{*R}_{x}[d][x][y]$ in the global
          memory (note that this array is transposed).
  \end{enumerate}
\item Increment $d$ if $d<D$, and go to step 3.
\end{enumerate}
In this step, $D$ arrays are stored in the global memory.

\subsection{Step4}
$L[y][x]$ and $R[y][x]$ have been transposed and stored as $L^{*}[x][y]$
and $R^{*}[x][y]$ in step2. By using these arrays, the brightness of the
pixels are compared efficiently along the $y$-axis. In this case, the
pixel data (for example $L^{*}[x][y]$) are compared horizontally
(parallel memory accesses are allowed only in this direction), and this
means that $L^{*}[x][y]$ are compared with $L^{*}[x][y+dy]$ ($dy =-W_y
,W_y$). The range of the similar pixels are stored in $W^{*L}_-[x][y]$
and $W^{*L}_+[x][y]$ for $L^{*}[x][y]$, and in $W^{*R}_-[x][y]$ and
$W^{*R}_+[x][y]$ for $R^{*}[x][y]$. $L^{*}[x][y]$ and $R^{*}[x][y]$ are
processed in parallel as shown in Fig.\ref{fig:pipeline}-$step4$, and
each SM processes $X/15$ columns. The $Y$ lines in each column are
assigned to $Y$ threads in the same way shown in Fig.\ref{fig:mapping}
though $x$ and $y$ are transposed.

\subsection{Step5}
In this step (Fig.\ref{fig:pipeline}-$step5$), $CA^{*L}_x[d][x][y]$ and
$CA^{*R}_x[d][x][y]$ are aggregated along the $y$-axis in parallel using
$W^{*L}_-[x][y]$, $W^{*L}_+[x][y]$, $W^{*R}_-[x][y]$ and $W^{*R}_+[x][y]$, and
then $D^L_{map}[y][x]$ and $D^R_{map}[y][x]$ (disparity maps when $L[y][x]$ and
$R[y][x]$ are used as the base) are also generated as follows.
\begin{enumerate}
\item Each SM processes $X/15$ columns.
\item $Y$ threads in each SM processes $Y$ pixels in each of the $X/15$ columns.
\item Each thread repeats the following steps (in the following, only the steps
  for the left image are shown).
  \begin{enumerate}
    \item Read $W^{*L}_-[x][y]$ and $W^{*L}_+[x][y]$ from the global memory.
      \item Set $d=0$.
    \item $Min[y]=${\it MAX\_VALUE} and $D_{map}[y]=0$.
    \item Calculate $CA^L(x,y,d)$ as follow.
      \begin{enumerate}
         \item Set $CA[y]=CA^{*L}[d][x][y]$.
           \item Add $CA^{*L}_x[d][x][y+dy]$ to $CA[y]$ starting from $dy=1$ to
             the position given by $W^{*L}_+[x][y]$.
           \item Add $CA^{*L}_x[d][x][y-dy]$ to $CA[y]$ starting from $dy=1$ to
             the position given by $W^{*L}_-[x][y]$.
        \end{enumerate}
    \item If $CA[y]<Min[y]$ then $Min[y]=CA[y]$ and $D_{map}[y]=d$.
    \item Increment $d$ if $d<D$, and go to step 3(c).
    \item Store $D_{map}[y]$ in $D^L_{map}[y][x]$ in the global memory (note that
      his array is re-transposed).
   \end{enumerate}
\end{enumerate}

\subsection{Step6}
In this step (Fig.\ref{fig:pipeline}-$step6$), $D^L_{map}[y][x]$ and
$D^R_{map}[y][x]$ are read from the global memory line by line, and the
condition for the GCP (described in Section 3.D) is checked. In this
step, each SM processes $Y/15$ lines. {\it Thread}$_x$ first accesses
$D^L_{map}[y][x]$, and then $D^R_{map}[y][+k]$ if
$D^L_{map}[y][x]=k$. $k$ is different for each thread, and bank conflict
happens in this step.

\subsection{Step7}
The median filter is applied using 15 SMs for the left image at
first. Then, the bilateral estimation is used to fill the non-GCPs. If
$D^L_{map}[y][x]$ is not a GCP, $thread_x$ scans $D^L_{map}[y][x]$ to
the $+/-$ direction of the $x$-axis in order, and finds two GCPs (the
GCPs closest on the left- and right-hand side). Then, the difference of
the disparities of the two GCPs is calculated. If the difference is
smaller than the threshold, the $D^L_{map}[y][x]$ is filled linearly
according to the disparities of the two GCPs. If the different is larger
than the threshold, the brightness of the two GCP are compared with the
target pixel, and the disparity of the target pixel is replaced by the
one which has similar brightness. Then, the improved disparity map
$D^{+L}_{map}[y][x]$ is stored in the global memory.

\subsection{Step8}
Finally, the disparity map $D^{+L}_{map}[y][x]$ is scaled up by using
the 15 SMs. In order to maintain a high accuracy, during the scaling-up
along the $x$-axis, the bilateral estimation method described above is
used again. On the other hand, the estimation along the $y$-axis is
applied linearly. Then, the final disparity map $D^{fL_{org}}[y][x]$ is
transferred back to the CPU.

\begin{table*}
\renewcommand{\arraystretch}{1.3}
\caption{Error rate when the cost aggregation range is changed (average error rate (\%))}
\label{error rate}
\centering
\begin{tabular}{lcccccc}
\hline \bfseries $W_L$$\backslash$$W_R$ &\bfseries $W_y=9$&\bfseries $W_y=11$ &\bfseries
$W_y=15$ &\bfseries $W_y=21$ &\bfseries $W_y=27$ &\bfseries $W_y=31$\\ \hline
$W_x=5$ &25.10&24.98&24.83&24.68&24.62&24.61\\ 
$W_x=9$&24.67&24.55&24.4&24.3&24.26&24.26\\
$W_x=21$&24.39&24.34&24.21&24.13&24.1&24.09\\ 
$W_x=41$&24.53&24.38&24.29&24.21&24.19&24.17\\
$W_x=61$&24.53&24.5&24.41&24.32&24.31&24.28\\
$W_x=141$&24.6&24.55&24.48&24.42&24.38&24.28\\
\hline
\end{tabular}
\end{table*}

\begin{table*}
\renewcommand{\arraystretch}{1.3}
\caption{Execution Time For The Middlebury Benchmark Set (ms)}
\label{runtime}
\centering
\begin{tabular}{lcccccccccccc}
\hline \bfseries Image &\bfseries Size &\bfseries Dmax &\bfseries SD
&\bfseries $W^{LR}_{\pm}$ &\bfseries $W^{*LR}_{\pm}$ &\bfseries $C+CA_x$ &\bfseries CA
&\bfseries CC &\bfseries Post&\bfseries SU&\bfseries Overall(GPU)&\bfseries FPS\\ \hline
Adirondack(H)&$1436\times{992}$&145&0.035&0.149&0.317&10.443&13.943&0.437&0.181&0.09&25.595&40\\ 
Pipes(H)&$1482\times{994}$&128&0.038&0.135&0.289&19.8&11.44&0.506&0.238&0.09&32.536&31\\ 
Vintage(H)&$1444\times{960}$&380&0.038&0.143&0.315&26.142&36.394&0.092&0.221&0.087&63.432&16\\
\hline
\end{tabular}
\begin{flushleft}
\textit{\textbf{Size}: $W^{LR}_{\pm}=21$, $W^{*LR}_{\pm}=31$, $TC=13$}
  \textit{\textbf{Dmax}: Maximum Disparity \textit{SD}:
    Scaling-Down. $W^{LR}_{\pm}$: Edge detection along the
    $x$-axis. $W^{*LR}_{\pm}$: Edge detection along the $y$-axis $C+CA_x$: cost
    calculation \& Aggregation along the $x$-axis. \textbf{CA}: Aggregation
    along the $y$-axis. \textbf{CC}: Cross\_check. \textbf{Post}: MedianFilter\&Bilateral estimation. \textbf{SU}: Scaling-up. \textbf{Overall}: The overall time taken on GPU.}
\end{flushleft}
\end{table*}

\begin{table*}
\renewcommand{\arraystretch}{1.3}
\caption{COMPARISON WITH HIGH-SPEED STEREO VISION SYSTEMS}
\label{comparison_systems}
\centering
\begin{tabular}{lcccccc}
\hline \bfseries System &\bfseries Size &\bfseries Dmax &\bfseries
Hardware &\bfseries Benchmark &\bfseries FPS &\bfseries MDE/s\\ \hline
RT-FPGA\cite{Global_FPGA} &$1920\times{1680}$ &60 &Kintex 7 &Middlebury v2 &30 &5806\\ 
FUZZY\cite{Fuzzy_logic} &$1280\times{1024}$ &15 &Cyclone II &Middlebury v2 &76 &1494\\
Low-Power\cite{Low_power} &$1024\times{768}$ &64 &Virtex-7 &Middlebury v2 &30 &1510\\ 
ETE\cite{End_end_GPU} &$1242\times{375}$ &256 &GTX TITAN X &KITTI 2015 &29 &3458\\
EmbeddedRT\cite{Embedded_GPU} &$640\times{480}$ &128 &Tegra X1 &KITTI 2012 &81 &3185\\
MassP\cite{Self-distributed} &$1440\times{720}$ &128 &GPU &Middlebury v3 &128 &3981\\
\bfseries Our system &$1436\times{992}$ &145 &GTX 780 Ti &Middlebury v3 &40 &7849\\
\hline
\end{tabular}
\end{table*}
\begin{figure*}
\centering
  \includegraphics[width=6in]{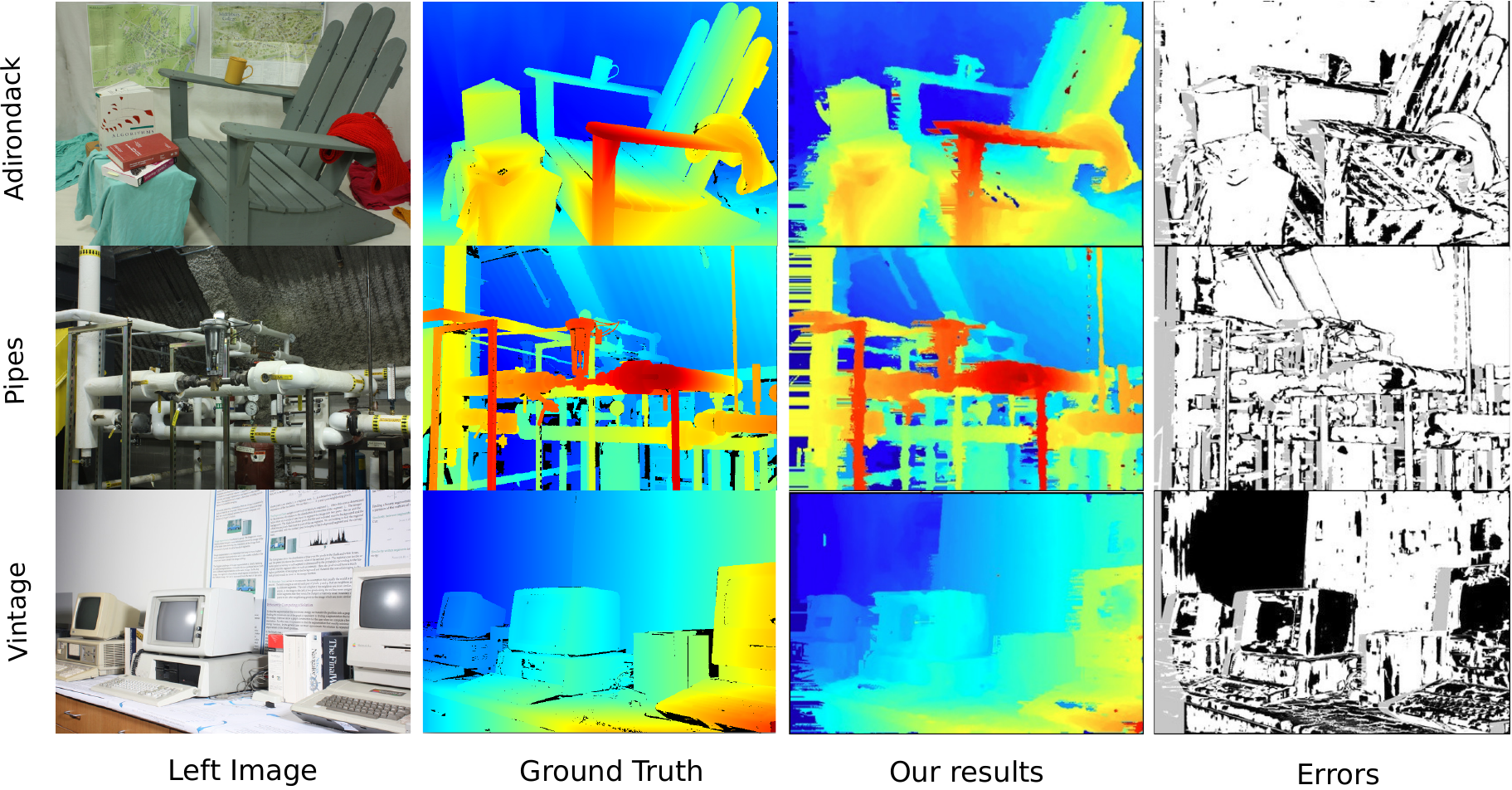}
    \caption{Processing results}
  \label{fig:results}
\end{figure*}

\section{experimental results}
We have implemented the algorithm on Nvidia GTX780Ti.  The error rate and the
processing speed are evaluated using Middlebury benchmark set \cite{Middlebury}.

In this evaluation, all parameters mentioned above affect the
performance of the accuracy. According to our tuning results, we first
set $\lambda_{AD}=0.3$, $\lambda_{MC}=2.3$ and $T=3$ to ensure a good
accuracy.
In the cost aggregation step, lower error rates can be expected by adding more
cost along the $x$- and $y$- axes, though it requires more computation time, and
makes the system slower. The maximum range of the cost aggregation $(W_x,W_y)$
can be changed when calculating $D^L_{map}$ and $D^R_{map}$. By changing them,
the different criteria are used for the left and right image, and the GCPs can
be more reliable. Table \ref{error rate} shows the error rate (\%) when the cost
aggregation range is changed. In Table \ref{error rate}, $W_{x}$ are the maximum
aggregation range along the $x$-axis for the left and right images, and $W_y$ is
the maximum aggregation range along the $y$-axis ($W_y$ is common to the left
and right images). As shown in Table \ref{error rate}, by enlarging $W_y$, the
error rates can be improved when $W_{x}$ is small. We have fixed $W_x=21$ and
$W_y=31$. To our best knowledge, the accuracy of our system is higher than other
real-time systems (like \cite{r200high}) which are listed in Middlebury
Benchmark \cite{Middlebury}. Additionally, we also compared our error rates with
that obtained using the original size image set, which are processed using
larger window ranges $W_x=41$ and $W_y=61$. According to our evaluation, the
error rate (Bad 2.0) of our system (24.09\%) is higher than that by the original
size images (32.82\%). One of the reasons is that we didn't tuned the parameters
$\lambda_{AD}$ and $\lambda_{MC}$ for the original image set. The other one is
that in the scaling down images, some information that makes the matching
difficult in the original size images, such as repetition of patterns and a
serious of similar pixels, are discarded, and better matching becomes possible.

The processing speed of our system is almost proportional to the window
size and the maximum disparity. Therefore, the computation time for
'Vintage' becomes the slowest. Table \ref{runtime} shows the processing
speed of our system and its details. We ignore the time for CPU-GPU data
transfers (less than 3\% of the total elapsed time) since it can be
overlapped with the computation. As shown in Table \ref{runtime}, most
of the computation time is used for the cost calculation and its
aggregation ($C+CA_x$ and $CA$). $W^{LR}_{\pm}$, $W^{*LR}_{\pm}$, $CC$,
$Post$, $SD$, $SU$ show the computation time for finding the cost
aggregation range along the $x$- and $y$- axes, cross checking,
post-processing and the image scaling. Unfortunately, for the 'Vintage'
set, its processing speed is 16fps due to the large disparity, and we
cannot achieve the real-time processing.

Table \ref{comparison_systems} compares the processing speed of our system with
other hardware systems. In Table \ref{comparison_systems}, all of the systems
achieved a real-time processing as shown in \textit{FPS} field, but their target
image size (\textit{Size}) and disparity range (\textit{Dmax}) are different.
According to the mega disparity evaluation per second (\textit{MDE/S}), it can
be noted that our system is much faster than other systems.

Fig.7 shows the results of our system for the two benchmark sets: Adirondack,
Pipes and Vintage.

\section{Conclusion}
In this paper, we have proposed a real-time stereo vision system for high
resolution images on GPU. Its processing speed is much faster than previous
one. At the same time, it also maintained a high accuracy. In our current
implementation, the processing speed is still limited by the access delay to the
global memory. To improve the processing speed so that the real-time processing
can be achieved slower GPUs is our main future work.

\end{document}